\documentclass[sigconf]{acmart}

\usepackage{microtype}
\usepackage{graphicx}
\usepackage{subfigure}
\usepackage{booktabs} 
\usepackage{amsmath}
\newtheorem{note}{Note}
\usepackage{hyperref}

\AtBeginDocument{%
  \providecommand\BibTeX{{%
    \normalfont B\kern-0.5em{\scshape i\kern-0.25em b}\kern-0.8em\TeX}}}

\copyrightyear{2021}   
\acmYear{2021}
\setcopyright{acmlicensed}\acmConference[ICAIF'21]{2nd ACM International
Conference on AI in Finance}{November 3--5, 2021}{Virtual Event, USA}
\acmBooktitle{2nd ACM International Conference on AI in Finance (ICAIF'21),
November 3--5, 2021, Virtual Event, USA} 
\acmPrice{15.00}
\acmDOI{10.1145/3490354.3494367} 
\acmISBN{978-1-4503-9148-1/21/11}

\begin{document}

\title{Adversarial Attacks on Machine Learning Systems for High-Frequency Trading}

\author{Micah Goldblum}
\authornote{Both authors contributed equally to this research.}
\email{goldblum@umd.edu}
\orcid{1234-5678-9012}
\author{Avi Schwarzschild}
\authornotemark[1]
\email{avi1@umd.edu}
\affiliation{%
  \institution{University of Maryland}
  \city{College Park}
  \state{Maryland}
  \country{USA}
}

\author{Ankit Patel}
\affiliation{%
  \institution{Rice University}
  \city{Houston}
  \state{Texas
  \country{USA}}}

\author{Tom Goldstein}
\affiliation{%
  \institution{University of Maryland}
  \city{College Park}
  \state{Maryland}
  \country{USA}
}

\renewcommand{\shortauthors}{Goldblum, et al.}

\begin{abstract}
Algorithmic trading systems are often completely automated, and deep learning is increasingly receiving attention in this domain.  Nonetheless, little is known about the robustness properties of these models.  We study valuation models for algorithmic trading from the perspective of adversarial machine learning.  We introduce new attacks specific to this domain with size constraints that minimize attack costs.  We further discuss how these attacks can be used as an analysis tool to study and evaluate the robustness properties of financial models. Finally, we investigate the feasibility of realistic adversarial attacks in which an adversarial trader fools automated trading systems into making inaccurate predictions.
\end{abstract}

\ccsdesc[500]{Computing methodologies~Machine learning algorithms}

\keywords{Machine learning, adversarial attack, finance, trading, HFT}

\maketitle
\section{Introduction}
\label{Introduction}

Machine learning serves an increasingly large role in financial applications.  Recent trends have seen finance professionals rely on automated machine learning systems for algorithmic trading \citep{almgren2001optimal}, as robo-advisers that allocate investments and rebalance portfolios \citep{moyer2015putting,meola2017robo}, in fraud detection systems that identify illicit transactions \citep{bhattacharyya2011data,abdallah2016fraud,randhawa2018credit}, for risk models that approve/deny loans \citep{binns2017fairness}, and in high-frequency trading (HFT) systems that make decisions on timescales that cannot be checked by humans \citep{hendershott2011does,angel2013fairness,arevalo2016high, avramovic2017we, klaus2017market,borovkova2019ensemble}. With the widespread use of such models, it is increasingly important to have tools for analyzing their security and reliability.

In the mainstream machine learning literature, it is widely known that many machine learning models are susceptible to {\em adversarial attacks} in which small but deliberately chosen perturbations to model inputs result in dramatic changes to model outputs \citep{kurakin2016adversarial}.   It has already been demonstrated that adversarial attacks can be used to change someone's identity in a facial recognition system \citep{sharif2016accessorize}, bypass copyright detection systems \citep{saadatpanah2019adversarial}, and interfere with object detectors \citep{eykholt2018robust}.
In this work, we investigate adversarial attacks on models for algorithmic and high-frequency trading.  

Trading bots historically rely on simple ML models and increasingly leverage the cutting-edge performance of neural networks \citep{borovkova2019ensemble, arevalo2016high}.  Security and reliability issues are particularly relevant to HFT systems, as actions are determined and executed in extremely short periods.
These short time horizons make it impossible for human intervention to prevent deleterious behavior, and it is known that unstable behaviors and security vulnerabilities have contributed to major market events (e.g., the 2010  ``flash crash''  \citep{kirilenko2017flash}).

We focus on adversarial attacks on the stock price prediction systems employed by trading bots.  In particular, we examine adversarial attacks from the following two perspectives.

\par \textbf{Adversarial analysis of reliability and stability.}   After a price prediction model has been trained, it is important to understand the reliability of the model and any unseen instabilities or volatile behaviors it may have.  A popular analysis method is backtesting, wherein one feeds recent historical stock market data into a model and examines its outputs.  However, models may have extreme behaviors and instabilities that arise during deployment that cannot be observed on historical data.  This is especially true when complex and uninterpretable neural models are used \citep{szegedy2013intriguing}.  Furthermore, market conditions can change rapidly, resulting in a domain shift that degrades model performance~\citep{de2018advances,pedersen2019efficiently}.  

We propose the use of adversarial attacks to reveal the most extreme behaviors a model may have.  Our adversarial attacks generate synthetic market conditions under which models behave poorly and can be used to determine whether instabilities exist in which small changes to input data result in extreme or undesirable behavior.  They are in part a tool to interrogate the reliability of a model before deployment or during postmortem analysis after a major event. 

\par \textbf{Adversarial attacks as a vulnerability.}  We assess the potential for adversarial attacks to be used maliciously to manipulate trading bots. One unique factor that contributes to the potential vulnerability of stock price models is how directly the data can be perturbed.  
In mainstream adversarial literature, attackers often cannot directly control model inputs; adversaries modify a physical object (\emph{e.g.}, a stop sign) with the hope that their perturbation remains adversarial after the object is imaged under unknown conditions.  In contrast, our adversary is blessed with the ability to directly control the input signal.  When order book data is used for price prediction, law-abiding traders and malicious actors alike receive identical market data from an exchange.  The adversary can perturb the order book by placing their own orders.  These adversarial orders quickly appear on the public exchange and are fed directly into victim models.  

At the same time, there are challenges to crafting attacks on order book data. The orders placed by an adversary interact with the market, so we develop a differentiable trading simulation which enables the adversary to anticipate how their own orders will affect the data seen by their victims.  This task cannot be solved by commercial backtesting software since this would not lend itself to automatic differentiation.  Additionally, an adversary's malicious orders must be bounded in their financial cost and detectability.  Moreover, the attacker cannot know the future of the stock market, and so they must rely on {\em universal} attacks that remain adversarial under a wide range of stock market behaviors. An adversary's knowledge of the victim model is also limited, thus we assess the effectiveness of these universal attacks across model architectures as well.

Interestingly, when deployed on historical data, we see that adversarial algorithms can {\em automatically} discover known manipulation strategies used by humans in ``spoofing''.  These attacks are known to be effective and problematic, spawning their ban in the Dodd-Frank Act of 2010 \citep{henning2018problem}.



%

\section{Background}
\subsection{Machine Learning for High-Frequency Trading}
\label{RelatedWork}
High-frequency trading bots can profit from extremely short-term arbitrage opportunities in which a stock price is predicted from immediately available information, or by reducing the market impact of a large transaction block by breaking it into many small transactions over a trading day \citep{hendershott2011does,angel2013fairness,klaus2017market,byrd2019intra}. These high-frequency trading systems contain a complex pipeline that includes price prediction and execution logic.  Price prediction models, or \emph{valuations}, are a fundamental part of any trading system since they forecast the future value of equities and quantify uncertainty. Execution logic consists of proprietary buy/sell rules that account for firm-specific factors including transaction costs, price-time priority of orders, latency, and regulatory constraints \citep{de2018advances}.    

Since the ability to anticipate trends is the core of algorithmic trading, we focus on price prediction and its vulnerability to attack.  Price predictors use various streams of time-series data to predict the future value of assets.  Traditional systems use linear models, such as autoregressive moving-average (ARMA) and support-vector machine (SVM), because models like these are fast for both training and inference \citep{beck2013empirical, kercheval2015modelling}.

\par As hardware capabilities increase, training and inference accelerate with advances in GPU, FPGA, and ASIC technologies.  Large models, like neural networks, have gathered interest as computations that were once intractable on a short timescale are becoming feasible \citep{dixon2018sequence}.  Recent works use LSTM and temporal convolutional networks to predict future prices with short time horizons \citep{borovkova2019ensemble, arevalo2016high}.

\par Common trading strategies, typical of technical analysis, rely on mean-reversion or momentum-following algorithms to identify an oversold or overbought asset.  Mean-reversion methods assume that fluctuations in stock prices will eventually relax to their corresponding long-term average, whereas momentum-following methods aim at identifying and following persistent trends in asset prices. 
Simple trading algorithms identify an indicator of either strategy and act when prices cross a predefined threshold, thus signaling for a potentially profitable trade over the oversold or overbought asset \citep{murphy1999technical,cohen2019trading}. We use this type of thresholding for our trading systems.  More details on our setup can be found in Section \ref{models}.

\par Historically, traders have engaged in \emph{spoofing} in which an adversary places orders on either side of the best price to fake supply or demand, often with the intent of canceling the orders before execution \citep{sar2017dodd}.  Spoofing is now illegal in the United States under the Dodd-Frank Act of 2010. These orders placed by a trader engaging in spoofing can be thought of as naive hand-crafted adversarial perturbations.  In this work, we use techniques from mathematical optimization to automate the construction of effective and efficient perturbations.  Our perturbations, crafted with optimization techniques, may be less detectable than handcrafted versions and thus may present a new problem for market regulators.

\subsection{Adversarial Attacks}
\label{AdversarialAttacks}
\par While it is known that neural networks are vulnerable to small perturbations to inputs that dramatically change a network's output, the most popular setting for studying these adversarial attacks to date is image classification.  Adversarial attacks in this domain perturb an image to fool a classifier while constraining the perturbation to be small in some norm.

While classical adversarial perturbations are designed for a single image, \emph{universal adversarial perturbations} are crafted to fool the classifier when applied to nearly any image \citep{moosavi2017universal,shafahi2018universal}. Other domains, such as natural language and graph structured data, have attracted the attention of adversarial attacks, but these attacks are impractical \citep{alzantot2018generating, dai2018adversarial}.  For example, in NLP, adversarial text may be nonsensical, and in graph structured data attacks are weak. 

\subsection{The Order Book}
\label{OrderbookData}

\par The \emph{order book} keeps track of buy orders and sell orders for a financial asset on an exchange.  We use \emph{limit} order book data from the Nasdaq Stock Exchange.  Limit orders are orders in which an agent either offers to buy shares of a stock at or below a specified price or offers to sell shares of a stock at or above a specified price.  Since agents can place, for example, buy orders at a low price, limit orders may not execute immediately and can remain in the order book for a long time.  When the best buy order price (the highest price at which someone will buy) matches or exceeds the best sell order price (the lowest price at which someone is willing to sell)  orders are filled and leave the book. 

At any given time, the order book contains prices and their corresponding sizes.  \emph{Size} refers to the total number of shares being offered or demanded at a particular price.  The size entries for a given price level may contain orders of many agents, and these orders are typically filled in a first-in-first-out manner.
%
 More complex rules can come into play, for example if orders require tie breaking, or if limit orders have stipulations on their execution.  A complete list of market practices and conduct standards can be found in the official Nasdaq equity rules \citep{nasdaq}. 

\section{Building Valuation Models}
\label{ProblemSetting}
In this section, we describe the simple valuation models that we attack. Our models digest a time series of size-weighted average price levels over a one minute interval, and make a prediction about a stock price ten seconds in the future.  The models used here were chosen for their simplicity -- they are meant to form a testbed for adversarial attacks and not to compete with the proprietary state of the art.  Still, we verify that our models learn from patterns in the data and outperform simple baselines.

\subsection{Data}

\par We use centisecond-resolution data from September and October 2019, and we only use the first hour of trading each day since this period contains the largest volume of activity.  We split one month of data per asset, using the first 16 days for training and the last four for testing.\footnote{A typical trading month is 20 days.}  We choose well-known stocks with high volatility relative to order book thickness: Ford (F), General Electric (GE), and Nvidia (NVDA). Order book data was furnished by LOBSTER, a financial data aggregator that makes historical order book data for Nasdaq assets available for academic use \citep{huang2011lobster}.

\par Consider that each row of the order book contains the ten best buy prices and their corresponding sizes, $\{(p_1^B, s_1^B), ..., (p_{10}^B, s_{10}^B)\}$, as well as the ten best sell prices and their corresponding sizes, $\{(p_1^S, s_1^S), ..., (p_{10}^S, s_{10}^S)\}$.  These rows are each a snapshot of the book at a particular time.  We process this data by creating the \emph{size-weighted average} (SWA), 
$$\text{SWA}([p_1^B, s_1^B, ..., p_{10}^B, s_{10}^B, p_1^S, s_1^S, ..., p_{10}^S, s_{10}^S]) $$ $$= \frac{\sum_{i=1}^{10}p_{i}^{B}s_i^B + \sum_{i=1}^{10}p_{i}^{S}s_i^S}{\sum_{i=1}^{10}s_{i}^{B} + \sum_{i=1}^{10}s_{i}^{S}},$$ 
for each row.  Movement in the size-weighted average may represent a shift in either or both price and size.  The SWA is a univariate surrogate for the price of an asset.

\par
\begin{figure}
    \centering
    \includegraphics[width=\columnwidth]{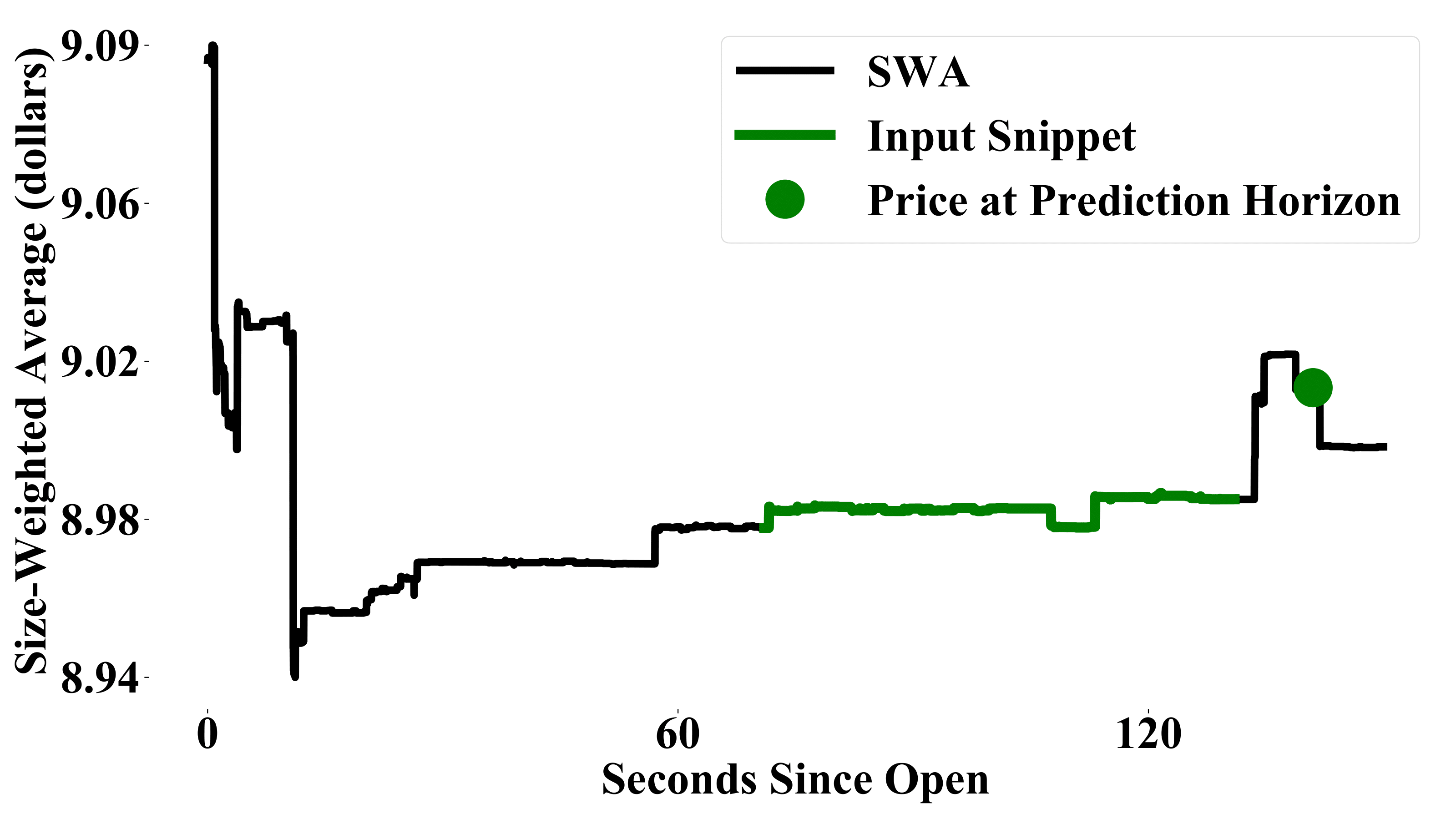}\\ \vspace{-10pt}
    \caption{This sample of the SWA curve for GE data shows an example of an input snippet. The label for this input is determined by the change in price from the right end of the green snippet to the green dot.}
    \label{fig:input_data}
\end{figure}

 After computing the SWA of the order book data, inputs to the models are $60$ second time-series of this one-dimensional data.  Since we use order book snapshots at intervals of $0.01$ seconds, a single input contains $6,000$ SWA entries. See Figure \ref{fig:input_data} for a visual depiction.

\par We focus on three-class classification models for concreteness, but the vulnerabilities we highlight may be artifacts of training on order book data in general. The model classifies an equity as likely to increase in price above a threshold, decrease below a threshold, or remain between the thresholds.  Thresholds are chosen to be symmetric around the origin, and so that the default class (``no significant change'') contains one third of all events, making the classification problem approximately balanced. 
%
%


\subsection{Training Valuation Models}
\label{models}

On each asset, we train a linear classifier, a multi-layer perceptron (MLP), and an LSTM \citep{lstm-schmidhuber}. The MLP models have 4 hidden layers, each of width 8,000. The LSTM models have 3 layers and a hidden-layer size of 100.

We train each model with cross-entropy loss over the three classes. Since our data can be sampled by taking any 60 second snippet from the first hour of any of the trading days in our training set, we randomly choose a batch of 60-second snippets from all of the training data and perform one iteration of SGD on this batch before resampling. These batches range in size from 2,000 to 3,000 data points depending on the model. Further details can be found in Appendix \ref{ModelTraining}.
\subsection{Comparison to Baselines}
\label{baselines}

We verify that our models are indeed learning from the data by checking that they exceed simple baselines. There are two natural baselines for this problem: a random guess and a strategy that always predicts the label most commonly associated with that stock in the training data.\footnote{Our classes are designed so that the middle class (little to no change in the SWA) accounts for one third of the training data. This leaves the `up' and `down' classes slightly off balance.}
The latter baseline always performed worse than a random guess on the test data, so the most natural performance baseline accuracy is $33.33\%$.  Table \ref{tab:rob_results} shows accuracy measurements which should be read with this in mind.  Note that price prediction is a difficult task, and profitable valuation models typically achieve performance only a few percentage points better than baselines \citep{byrd2019intra}.

 Since we have a large quantity of data, we randomly sample 10,000 snippets from the test set in order to measure accuracy.  We report confidence intervals in Table \ref{tab:clean_results}.  All test accuracy measurements in this work have standard error bounded above by $100\sqrt{(0.5)(1-0.5)/10000}\%=0.5\%$, so confidence intervals are small, and performance differences are statistically significant.
 
 \begin{table*}[ht!]
\centering
\caption{Accuracy of each model on test data with confidence intervals of radius one standard error. Note that all models are performing pattern recognition as described in Section \ref{baselines}. Furthermore, the neural networks consistently outperform the linear models.}
	\label{tab:clean_results}
	\begin{tabular}{lr}
	    \multicolumn{2}{c}{Ford (F)} \\
		Model &  Test Accuracy(\%) \\ \midrule
        Linear  &  $34.66 (\pm0.48)$  \\
        MLP &  $36.56 (\pm0.48)$ \\ 
        LSTM &  $36.40 (\pm0.48)$  \\            
		\bottomrule
	\end{tabular} ~~
	\begin{tabular}{lr}
	    \multicolumn{2}{c}{General Electric (GE)} \\
		Model & Test Accuracy (\%) \\ \midrule
        Linear  &  $36.41 (\pm0.48)$ \\
        MLP &  $36.87(\pm0.48)$ \\ 
        LSTM &   $35.65 (\pm0.48)$  \\            
		\bottomrule
	\end{tabular}~~
	\begin{tabular}{lr}
	    \multicolumn{2}{c}{Nvidia (NVDA)} \\
		Model & Test Accuracy (\%) \\ \midrule
        Linear  &  $34.15 (\pm0.47)$\\
        MLP &  $35.69 (\pm0.48)$\\ 
        LSTM &  $37.11 (\pm0.48)$  \\            
		\bottomrule
	\end{tabular}
\end{table*}

\section{Analyzing the Vulnerabilities of Trading Models}
\label{Analyzing}

In this section, we develop a basic adversarial attack, and we use it to study the robustness properties of trading models as well as to compare the sensitivities of different networks.  One might think that because these models are trained on noisy data, they are insensitive to small perturbations.  We demonstrate that the networks are indeed vulnerable.

\subsection{Rules for Perturbing the Order Book}
\label{PropagatingPerturbations}
A number of unique complexities arise when crafting adversarial attacks on order book data.  We consider an adversary that may place orders at the ten best buy and sell prices (orders at extreme price levels are typically discarded by analysts as they are unlikely to be executed).  Equivalently, we consider perturbations to the size entries, but only to whole numbers, in the raw order book data.  

Note that using the SWA as input and only perturbing size entries prevents the adversary from making too great an impact on the data seen by the model.  Since the SWA lacks the informational content of raw order book data, an attacker, without changing the price entries, can at most perturb the SWA to the maximum or minimum price in the ten best buy/sell orders at a given time-stamp.  We will see that this small range gives the attacker enough freedom to inflict damage.


We now describe a simple differentiable trading simulation which enables the adversary to anticipate the impact of their orders on the order book snapshots digested by their victims. Both the algorithm and its implementation need to be carefully considered to ensure that they are reasonably fast and differentialble.

\par \textbf{Propagating orders through the book.} Consider the sequence of order book snapshots used for size-weighted average, $\{\mathbf{x}_i\}$, and the sequence of adversarial orders (on a per-row basis), $\{\mathbf{a}_i\}$.  We cannot simply add these two quantities to compute the perturbed order book since the adversarial orders may remain on the book after being placed.  On the other hand, we cannot blindly propagate the orders to the end of the snippet because transactions may occur in which adversarial orders are executed, removed from the book, and therefore excluded from subsequent snapshots.  Thus, denote the propagation function that accumulates adversarial orders, accounts for transactions, and returns the sequence of perturbations to the snapshots $\{\delta_i\}$, by $p_{\{\mathbf{x}_i\}}\left(\{\mathbf{a}_i\}\right)$.  The function $p_{\{\mathbf{x}_i\}}$ takes as input a sequence of orders, which indicates an integer number of shares to add to the order book at a particular time (or row in the raw data). This quantity is added to all subsequent rows in the data that include orders at that price, until any transaction occurs at that price, at which point our model considers the order filled. We make this assumption about limit order book execution as is done by commercial backtesting software \cite{multicharts}. The output is a perturbation $\{\delta_i\}$. Finally, the input to a model is $\text{SWA}(\{\mathbf{x}_i+\delta_i\})$.  The propagation function can be thought of as a basic differentiable backtesting engine that we will use to compute gradients through the adversary's interactions with the market.

The following complexity arises when implementing propagation.  The order book is organized by price level, or distance from the best price.  Since the best price can change, this format does not lend itself to propagating orders, since each order is placed at a fixed dollar value. To add size at a particular price then requires that different entries in each row must be modified.  If price levels do not shift in a snippet, then one column of data organized by price level corresponds to one column of data organized by absolute price, so an adversary's orders can be added column-wise to the latter representation.  However, price levels shift frequently, so translating between these two views of the order book data in a fast, parallelized fashion is non-trivial.  At the same time, the adversary will compute numerous gradient steps during an attack and must translate between the two data representations during each step.  This translation task must be handled efficiently by the propagation function.

\subsection{Quantifying Perturbation Size}
\label{Efficiency}

\par In order to determine just how sensitive our models are, we must quantify the size of the perturbations being placed.  We consider three possible metrics: cost, capital required, and relative size.  

We compute {\em cost} to the attacker as the change in the total value of his or her assets from the start of the snippet to the end. Since assets include both cash and stocks, cost accounts for changes in the price of stocks purchased when a buy order is filled (and we exclude transaction fees).  We operate under the assumption that all attack orders are transacted when other orders at that price are transacted. {\em  Capital required} is the hypothetical total cost of executing all orders placed by the attacker.  This quantity is the total dollar value of orders in the perturbation. An agent would need to have this much capital to place the orders. {\em Relative size} of a perturbation refers to the size (number of shares) of the adversary's propagated perturbation as a percentage of total size on the book during the snippet.

Since transactions occur infrequently at this time-scale, and prices go up roughly as much as they go down, the average cost across attacks is less than $\$1.00$ for each asset-model combination.  Therefore, our algorithms limit perturbation size in terms of capital required. We denote the capital required for an attack $\{\mathbf{a}_i\}$ as $C(\{\mathbf{a}_i\}).$  Additionally, we measure cost and relative size in our experiments.

\begin{table*}
\centering
	\caption{Model performance on clean and perturbed inputs}
	\label{tab:rob_results}
	\begin{tabular}{llrrrrrr}
		&Model & $\mathcal{A}_\text{test}(\%)$  & $\mathcal{A}_\text{rand} (\%)$&$\mathcal{A}_\text{adv} (\%)$ & Capital (\$) &  Size  (\%) \\ \midrule
  F &      Linear & $34.66$ & $-0.27$ & $-18.00$ &  $26,808$ & $0.4$\\
        &MLP & $36.56$ & $-3.44$ & $-29.34$ &  $24,055$ & $0.4$ \\ 
        &LSTM &  $36.40$ & $-0.86$ & $-14.64$ & $10,340$ & $<0.1$ & \\            
		\bottomrule
		\\
		&Model & $\mathcal{A}_\text{test} (\%)$ & $\mathcal{A}_\text{rand} (\%)$ &$\mathcal{A}_\text{adv} (\%)$ &Capital (\$) & Size  (\%)  \\ \midrule
       GE &Linear  & $36.41$ & $-1.38$ & $-16.88$ &  $34,513$ & $0.6$\\
        &MLP & $36.87$ & $-3.22$ & $-27.00$ &  $ 23,853$ & $0.4$\\ 
       & LSTM &  $35.65$ & $-2.64$ & $-15.06$ & $12,580$ & $<0.1$ & \\            
		\bottomrule
		\\
	&	Model & $\mathcal{A}_\text{test} (\%)$ & $\mathcal{A}_\text{rand} (\%)$ & $\mathcal{A}_\text{adv} (\%)$& Capital (\$)& Size  (\%)  \\ \midrule
      NVDA & Linear & $34.15$ & $+3.32$ & $-5.17$ &  $53,805$ & $2.5$ \\
       & MLP & $35.69$ & $-1.30$ & $-8.09$ &  $49,556$ & $2.4$ \\ 
       & LSTM &  $37.11$ & $-1.23$ & $-1.55$ & $5,327$ & $<0.1$ & \\           
		\bottomrule
	\end{tabular}
\end{table*}

\subsection{Robustness to Random Noise}
\label{Random}

\par Before describing our attack, we establish a baseline by randomly inserting orders instead of using optimization. We propagate the perturbations through the order book with a propagation function $p$. We compare our attacks to this baseline to demonstrate that the optimization process is actually responsible for impairing classification.  In fact, random attacks with a high budget barely impair classification at all. See Table \ref{tab:rob_results} where we denote test accuracy on the test data, changes to test accuracy resulting from random perturbations, and changes to test accuracy resulting from adversarial perturbations by $\mathcal{A}_\text{test}$, $\mathcal{A}_\text{rand}$, and $\mathcal{A}_\text{adv}$, respectively. We also report the average capital the adversary must have before making an effective perturbation, and the average relative size of successful attacks, both of which are computed as described in Section \ref{Efficiency}. The average cost to the attacker per asset is less than $\$1.00$. All random attacks have a budget over $\$$2,000,000, a sum far higher than any optimization-based attack we consider.

\subsection{Crafting Gradient-Based Attacks}
\label{Untargeted}

\par We now introduce a simple adversarial attack for studying model sensitivity.  This attack is untargeted, meaning that it degrades model performance by finding  a small perturbation that causes misclassification with any incorrect label.  

Consider a model $f$ and a sequence of snapshots $\{\textbf{x}_i\}$ with a label $y$.  Our attacker solves the maximization problem, 
$$
\max_{\{\mathbf{a}_i\}} \mathcal{L}\big(f(\text{SWA}(\{\mathbf{x}_i+\delta_i\})), y\big),
$$
 with capital constraint $C(\{\mathbf{a}_i\}) \leq c$, where $\{\delta_i\}=p(\{\mathbf{a}_i\})$.  The adversary cannot perturb price in the snapshot, and all size entries must be integer-valued.  To this end, we perform gradient ascent on the cross-entropy loss, $\mathcal{L},$ using a random learning rate $\alpha^k \sim \mathcal{U}(0,\alpha_0)$ and update rule,
$$
\{\mathbf{a}_i\}^k \leftarrow \{\mathbf{a}_i\}^{k-1} + \alpha^{k} \nabla \mathcal{L}\big(f(\text{SWA}(\{\mathbf{x}_i+\delta_i\})), y\big).
$$
The attacker runs through this iterative method with a small step size until either the model no longer predicts the correct label, $y$, or $C(R_{r}(\{\mathbf{a}_i\}))$ exceeds the capital constraint, $c$, where $R_{r} = \lfloor
\mathbf{x}+r \rfloor$ is a floor operator that rounds size entry $x$ to the greatest integer less than or equal to $x+r$ (i.e. $R_{0.3}(0.8)=1$).  When the attack terminates, we return the adversarial perturbation, $R_{r}(\{\mathbf{a}_i\})$.  We choose $c$ and $r$ to be $\$$100,000 and $0.95$, respectively, in our experiments.  See Section \ref{sec:attack_params} for a discussion on choice of attack hyperparameters.

\begin{note}
In a standard PGD attack, attacks are randomly initialized in an $\ell_{\infty}$ ball, and perturbations are projected onto the ball each iteration.  In contrast to $\ell_{\infty}$--bounded attacks, there is no natural distribution for random initialization within the set, $\{ \{\mathbf{a}_i\}: C(\mathbf{a}_i)\leq c\}$. Moreover, there is no unique projection onto this set. Thus, we instead opt for a randomized learning rate in order to inject randomness into the optimization procedure \citep{chiang2019witchcraft}.
\end{note}

\par Table \ref{tab:rob_results} depicts the effect of adversarial orders crafted in this manner on price prediction.  Empirically, these trading models are not robust to low-budget perturbations.  Moreover,  gradient-based attacks are far more effective at fooling our models than randomly placed orders with far higher budget, indicating that our valuation models are robust to noise but not to cleverly placed orders.  MLPs, while generally of higher natural accuracy than linear models, are also significantly less robust.  In all experiments, MLPs achieve lower robust accuracy, even with much smaller perturbations.  On the other hand, LSTM models seemingly employ gradient masking as they are far more difficult to attack using gradient information.  To further confirm the gradient masking hypothesis, we see below in Section \ref{Universal} that LSTM models are the most easily fooled models by transferred universal perturbations.

\subsection{The Effect of High Price-to-Size Ratio}
\label{Quantizing}

\par We find that valuation models for equities with high price per share and low trading volume were difficult to attack.  For such stocks, the increments by which an adversary can perturb the SWA are coarse, so gradient-based methods with rounding are ineffective.  For example, in the data we use, Nvidia stock costs  around $\$180$/share compared to about $\$9$/share for Ford stock during the date range we considered.  This fact, combined with roughly one fifth of the trading volume, makes it difficult to perform low cost attacks, or attacks with low relative size, on Nvidia. As a result of this quantization effect, we are almost unable to impair models on Nvidia.  This observation leads us to believe that price prediction on assets with this property is more robust in general (at least using the class of SWA-based models studies in our setting).  In the ensuing sections, we focus on Ford and General Electric stocks, which have relatively higher volume and lower cost per share.

\begin{figure*}[ht!]
\centering
\vspace{-14pt}
\includegraphics[width=0.8\columnwidth, trim=0.5cm 0.5cm 0.5cm 0.5cm, clip]{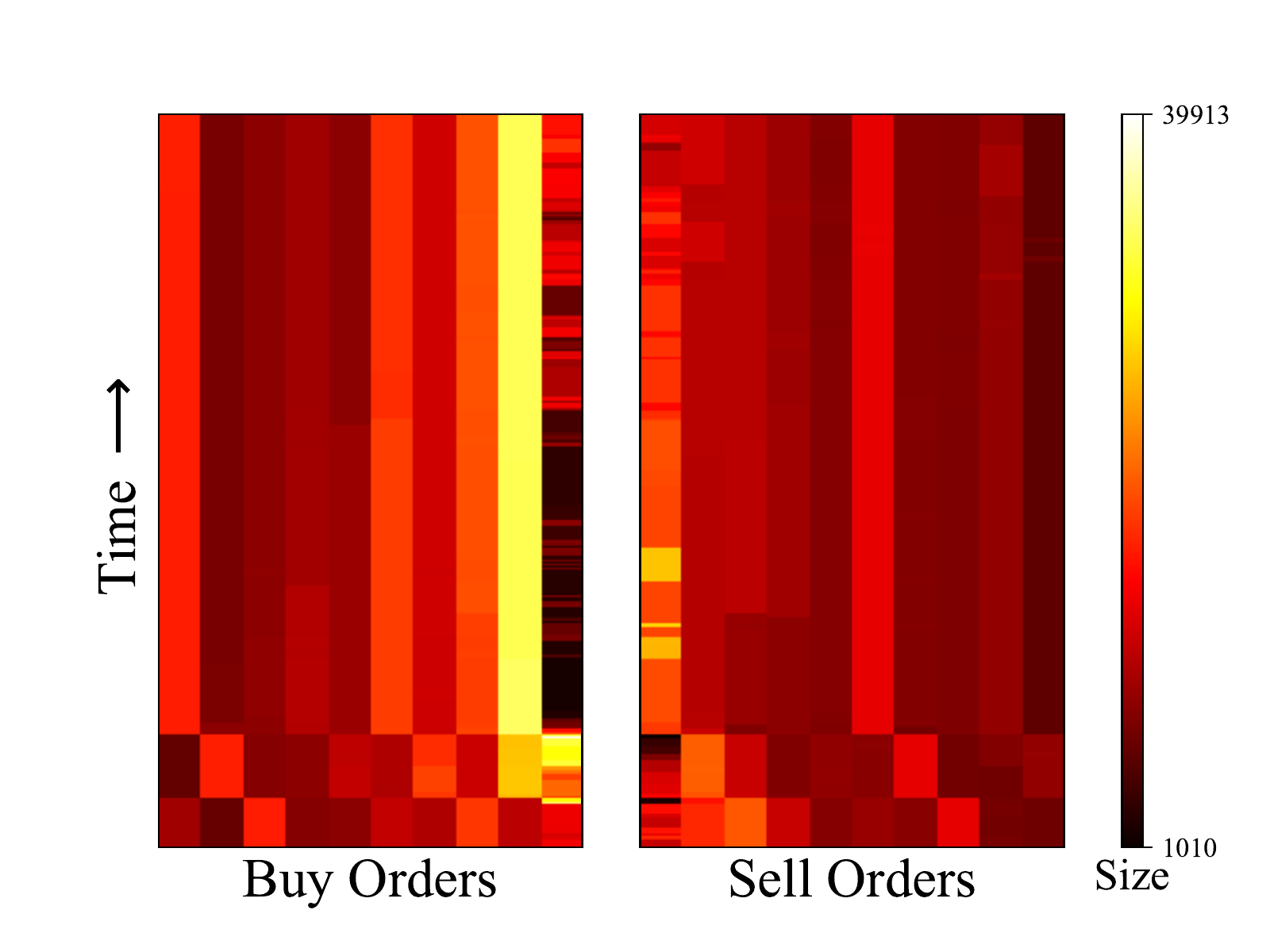}
\includegraphics[width=0.8\columnwidth, trim=0.5cm 0.5cm 0.5cm 0.5cm, clip]{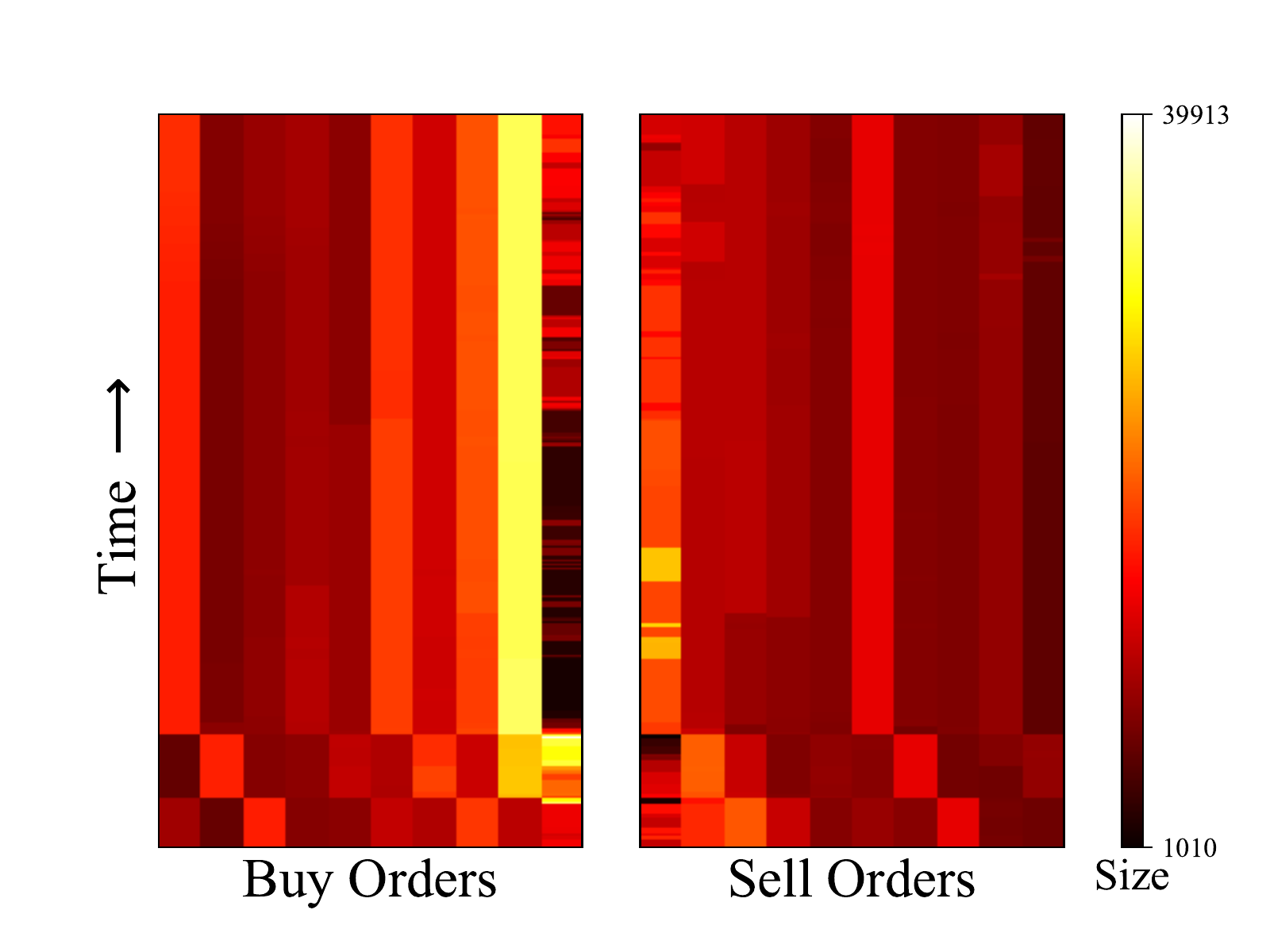}  \\
\includegraphics[width=0.8\columnwidth, trim=0.5cm 0.5cm 0.5cm 0.5cm, clip]{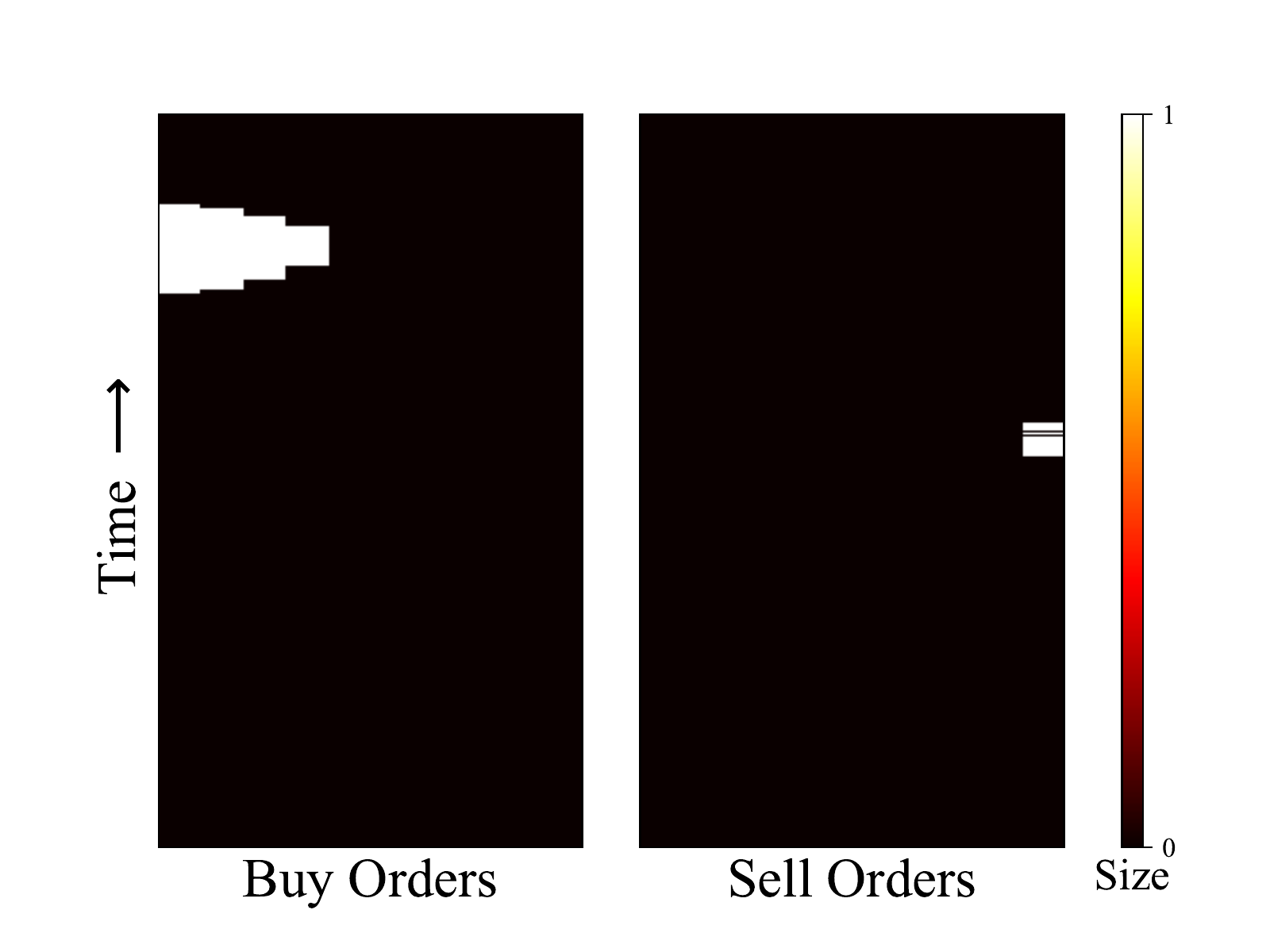}
\includegraphics[width=0.8\columnwidth, trim=0.5cm 0.5cm 0.5cm 0.5cm, clip]{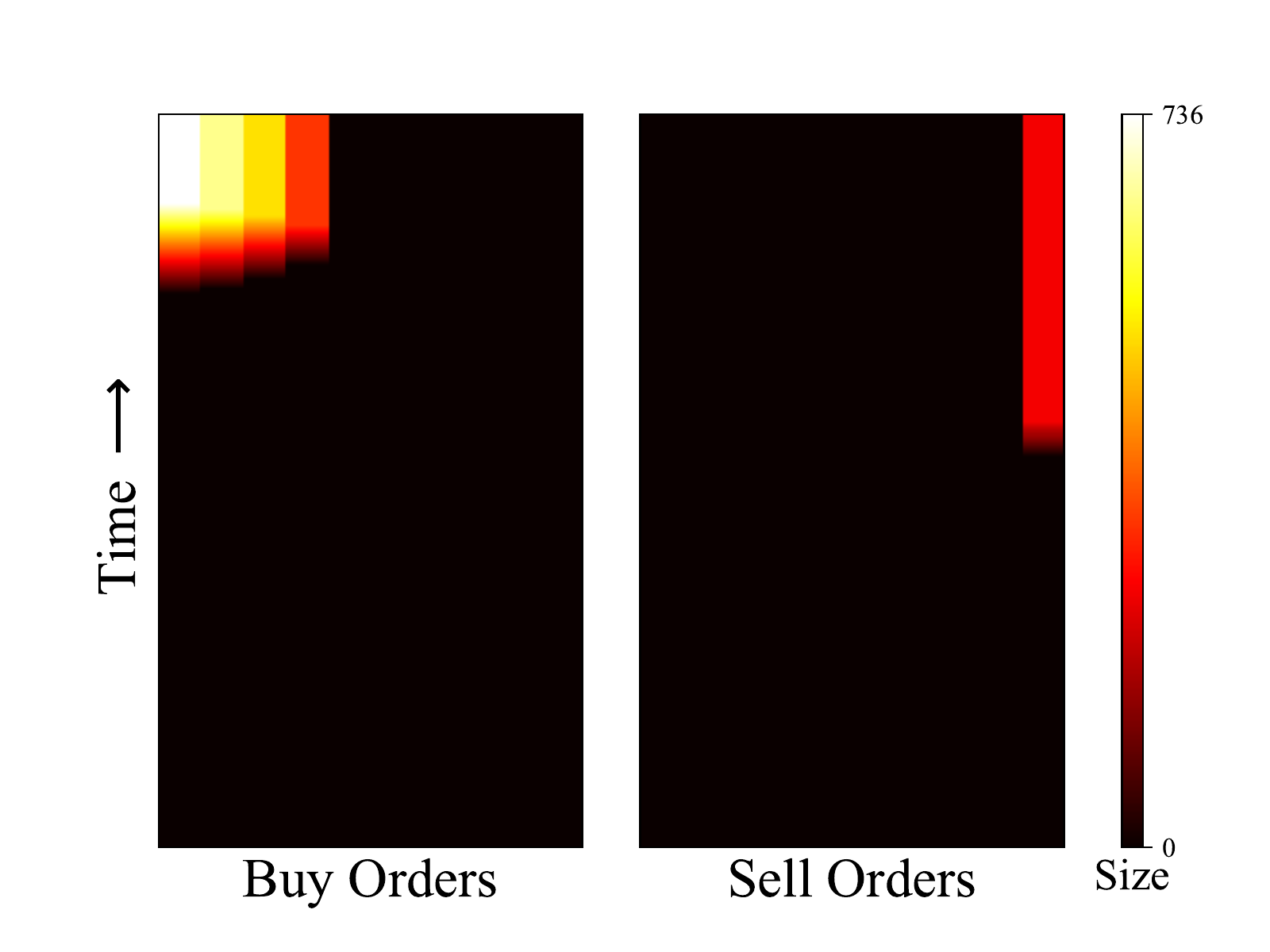}
    \caption{The unperturbed and perturbed order books (top left and top right, respectively). The bottom two images show the unpropagated and propagated adversarial perturbations (left and right, respectively). In this instance, the addition of the bottom right image to the clean signal in the top left yields the perturbed input in the top right. Note that the difference between the perturbed and unperturbed signals is not noticeable here, where the model was fooled into incorrectly classifying the SWA as going up.}
\label{fig:4panel}
\end{figure*}

\section{Universal Transfer Attack:  A Feasible Threat Model}
\label{Universal}

\par  The adversarial attacks discussed above provide a useful basis for analyzing model robustness but do not pose any real security threat because of three unrealistic assumptions. First, the attacker is performing an untargeted attack, meaning that the attacker knows the class that a snippet belongs to (i.e., whether it will go up, down, or remain), and is only trying to cause the prediction to change.  Second, attacks are crafted while looking at every row of the order book simultaneously.  For this reason, the attacker may use knowledge about the end of a snippet when deciding what orders to place in the book at the beginning of the snippet.  Third, the attacker must know their victim's architecture and parameters.  In short, the strategies displayed above require the attacker to know the future and their opponent's proprietary technology -- an entirely unrealistic scenario.   

In this section, we consider making {\em universal perturbations}. We craft a single attack that works on a large number of historical training snippets with the hope that it transfers to unseen testing snippets.  These attacks can be deployed in real time without knowing the future.   Furthermore, we use targeted attacks, so that the attacker may anticipate the behavior of the victim.  In order to add even more realism, we use transfer attacks in which the attacker does not know the parameters of the model they are attacking.  In this situation, we train surrogate models and use universal perturbations generated on the surrogates to attack the victim.

\par We craft universal perturbations by solving
 $$\min_{\mathbf{a}_i} \sum_{i}\mathcal{L}\big(f(\text{SWA}(\{\mathbf{x}_i+\delta_i\}), y_t\big)+\gamma T(\{\delta_i\}),$$
where $\mathcal{L}$ is cross-entropy loss, $\{\delta_i\}=p_{\{\mathbf{x}_i\}}(\{\mathbf{a}_i\})$, $\{x_i\}$ is a set of order book snapshots in the training data, $y_t$ is a target label, and $T(\{\delta_i\})$ is a penalty term which may be used to encourage the perturbation to require, for example, less size relative to the size on the book.  The resulting perturbation $\{\mathbf{a}_i\}$ represents the size of orders placed by the adversary at particular price levels and time stamps.  That is, the adversary must only consider how many levels a price is from the best offer rather than specific prices or dollar amounts.  We solve this problem with SGD by sampling training data and minimizing loss.  See Section \ref{sec:attack_params} below for hyperparameters.  We apply this pre-computed sequence of adversarial orders to test data by propagating orders through the order book for each individual input sequence and adding to the original snippet. We measure and report the success rate of labeling correctly classified inputs from outside the target class with the target label, and we refer to this as \emph{fooling} the model.

\par We find that universal adversarial perturbations in this setting work well. In particular, we were able to find universal perturbations that are small relative to the order book and yet fool the model a significant percentage of the time.  Moreover, we find that these attacks transfer between models and that our relative size penalty is effective at reducing the budget of an attacker while maintaining a high fool rate.  See Table \ref{tab:transfer_results} for targeted universal perturbation transfer attack results. We denote the universal perturbations computed on the linear model and on the MLP by $\mathcal{U}_\text{linear}$ and $\mathcal{U}_\text{MLP}$, respectively. We craft universal adversarial perturbations on a linear model and an MLP and assess the performance of three models under these attacks. Both of these attacks were computed with a penalty on relative size.

\par These targeted universal perturbations are prototypical patterns for convincing a model to predict a particular SWA movement.  For example, to force the victim to predict a downwards trend, the adversary places sell orders followed by buy orders and finally more sell orders to feign downwards momentum in the SWA. See the visualization of a universal perturbation in Figure \ref{fig:univ}, which is helpful for interpreting why our models make the decisions they do. 
\begin{table}[h!]
\centering
\caption{Model performance on universal attacks (Ford data)}
	\label{tab:transfer_results}
	\begin{tabular}{lrrrr}
	    \toprule
		 &\multicolumn{2}{c}{ $\mathcal{U}_\text{linear}$ } & \multicolumn{2}{c}{ $\mathcal{U}_\text{MLP}$ }\\
		 \cmidrule{2-3} \cmidrule{4-5} 
		Model &   Fooled  &  Size  &  Fooled &  Size  \\ 
		\midrule
        Linear & $9.5\%$ & $1.0\%$ & $11.90\%$ & $0.8\%$\\
        MLP & $23.75\%$ & $1.1\%$& $36.10\%$ & $0.8\%$\\
        LSTM & $31.21\%$ & $0.9\%$ & $40.46\%$ & $0.8\%$ \\
		\bottomrule
	\end{tabular}
\vspace{-12pt}
\end{table}

\label{Patterns}

\section{Patterns in Adversarial Orders}

\begin{figure}[h!]
\vspace{-12pt}
    \includegraphics[width=\columnwidth, trim=0.5cm 0.5cm 0.5cm 0.5cm, clip]{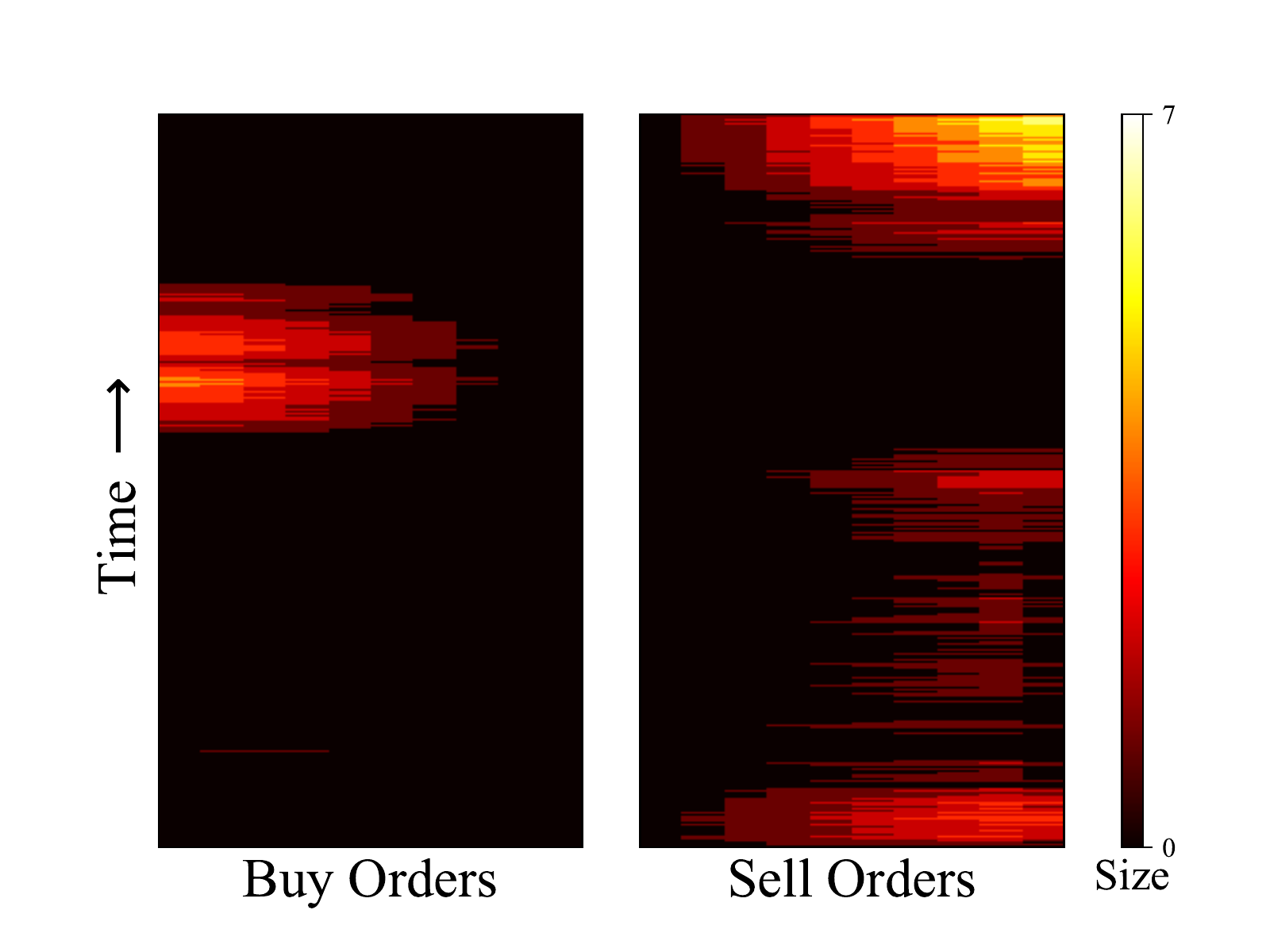}\\
    \includegraphics[width=\columnwidth, trim=0.5cm 0.5cm 0.5cm 0.5cm, clip]{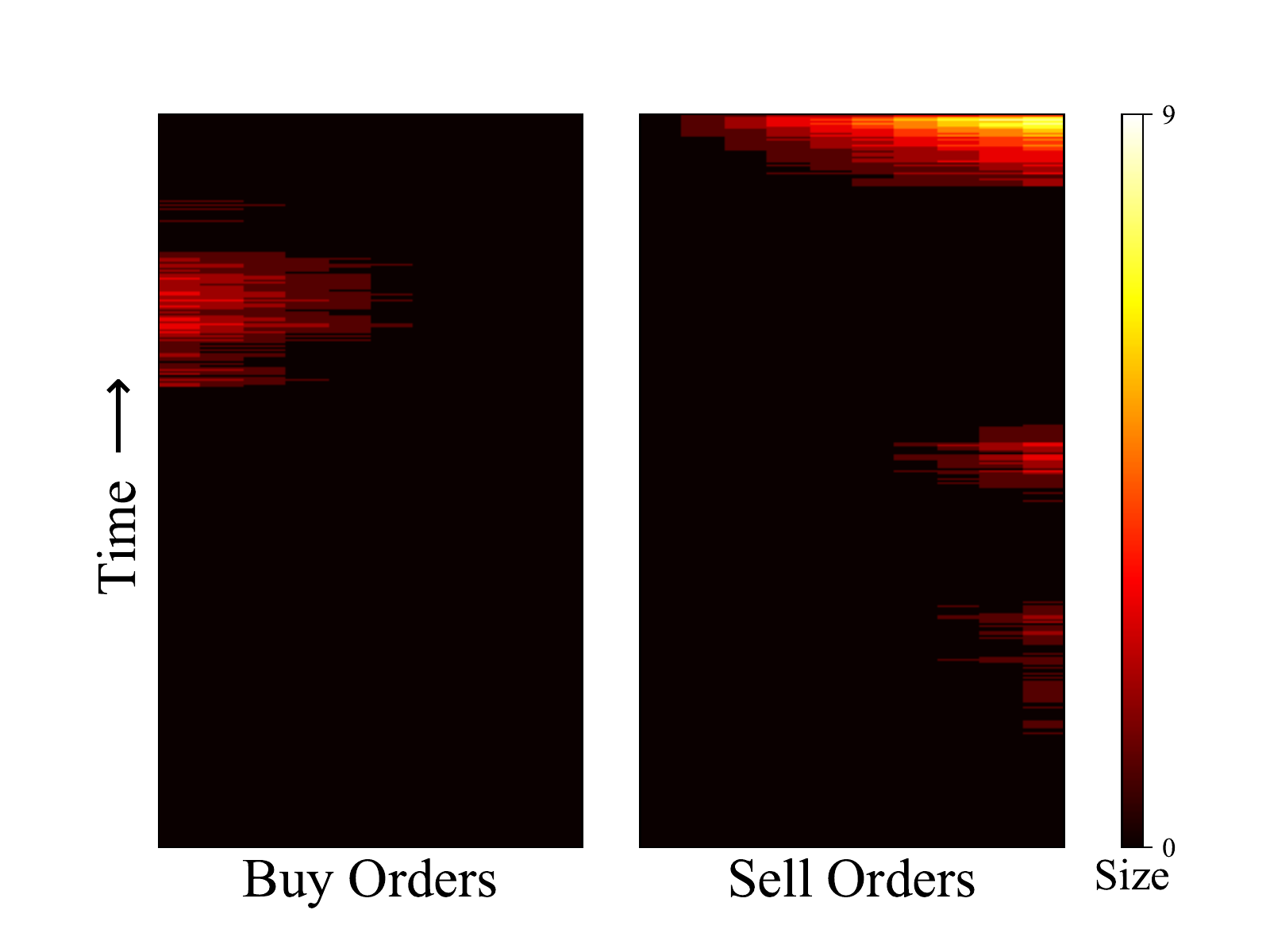}\\ \vspace{-10pt}
    \caption{Two targeted universal adversarial attacks computed on the same model (MLP trained on Ford data).}
\label{fig:univ}
\end{figure}

We observe patterns in adversarial attacks that can help explain the behavior of classifiers.  The non-universal attacks on particular inputs highlight the vulnerability of the valuation models. For example, in Figure \ref{fig:4panel}, the perturbation is so small compared to the size on the book that a human would not distinguish the attacked signal from the clean one.  The attack, however, is concentrated on the fringes of the order book, much like a spoofing attack.  Figure \ref{fig:univ} shows a case in which a universal adversary has learned an interpretable perturbation in which it creates a local minimum in the size-weighted average by alternately placing perturbations on opposite sides of the book. The perturbation on the top of Figure \ref{fig:univ} was computed without constraint, and when transferred to test data and a different model, it fooled the victim on 157 inputs (out of 341 correctly classified) and accounts for a relative size on the book of $3.8\%$. The perturbation on the bottom is the result of the same process with an added penalty on relative size. The penalized attack shows a sparser perturbation with relative size $0.9\%$ while causing misclassification of almost the same number of inputs (123).

Universal perturbations bring to light two important features of these price predictors. The first is that targeted universal perturbations, which are generally expected to be less effective, have a major impact on the performance of these models. The models are vulnerable even to these weak attacks.  Second, the targeted universal attacks expose interpretable model behavior.  In Figure \ref{fig:univ}, we see that the attacker creates local extrema in the order book which cause the model to anticipate mean reversion.  

\section{Hyperparameters and Implementation Details}
\label{ModelTraining}

\subsection{Training}
For each asset, we train linear predictors, MLPs, and LSTMs. The linear models are trained for 5,000 iterations of SGD with a batch size of 2,500 and an initial learning rate of 0.5. The learning rate is then decreased by a factor of 0.5 after iterations 50, 500, 1,000, 2,000, 3,000, 4,000 and 4,500. We conducted an architecture search, which led to MLP models of 4 hidden layers, each of 8,000 neurons. The MLPs are trained for 3,000 iterations of SGD with a batch size of 3,000 and an initial learning rate of 0.01. The learning rate decreases by a factor of 0.5 after iterations 10, 20, 100, 200, 400, 500, 1,000, 2,000, and 2,500. Our architecture search for LSTM models led to LSTMs with 3 layers and a hidden size of 100 neurons. The LSTMs are trained for 50 iterations of ADAM with a batch size of 2,000 and an initial learning rate of 0.01. The learning rate decreases by a factor of 0.5 after iterations 25 and 40.

\subsection{Attacks}
\label{sec:attack_params} 
When conducting non-universal attacks, we use $200$ steps and $\alpha_0 = 40$, so that $\alpha \sim \mathcal{U}(0,40)$.  The attack terminates on a given sample as soon as the sample is misclassified, so that we rarely actually run through all $200$ steps.  The run-time of these attacks is dominated by the gradient computation, and the worst-case time complexity of $200$ gradient computations per sample is rarely achieved.

For universal attacks, we adopt a similar procedure to \cite{moosavi2017universal}.  We start by initializing an attack, $\{\mathbf{a}_i\}$ to be zeros.  Then, we sample batches of $50$ training samples and perform $5$ gradient descent steps on each one with a random learning rate, $\alpha\sim\mathcal{U}(0,10)$.  Next, we average the resulting steps and add $1.5$ times this average to $\{\mathbf{a}_i\}$, clipping all negative values.  We use $200$ batches to create one universal adversarial perturbation.  In order to create sparse perturbations, in some experiments, we use a penalty term which is $5$ times the relative size of the perturbation.

In order to propagate perturbations through the order book in a differentiable way, we must store data in two forms: raw order book snapshots and a form which stores size entries corresponding to the same price together.  This doubles the space complexity but allows us to parallelize propagation.  During propagation, each entry is touched exactly twice, so this process is $\mathcal{O}(n)$.

\section{Discussion}
\label{Discussion}

This work introduces adversarial attacks to financial models both as a method for understanding model sensitivities and as a potential threat.  As seen in other applications of adversarial machine learning, there are robustness trade-offs, and there is no free lunch.  While neural network models perform better at pattern recognition in this setting than traditional linear valuation models, we find them to be less robust.  We further notice that the same adversarial patterns that fool one model also fool others and that these patterns are highly interpretable to humans.  The transferability of these attacks, combined with their ability to be effective with a small attack budget, suggests that they could possibly be leveraged by a malicious agent possessing only limited knowledge. 

\section*{Acknowledgements}
This work was supported by the AFOSR MURI program, DARPA GARD, and JP Morgan Chase.

\bibliographystyle{ACM-Reference-Format}
\bibliography{main}

\end{document}